\documentclass[letterpaper, 10 pt, conference]{ieeeconf}  

\IEEEoverridecommandlockouts                              

\usepackage{graphics} 
\usepackage{graphicx}
\usepackage{times} 
\usepackage{amsmath} 
\usepackage{amssymb}  
\usepackage{cite}
\usepackage{tabularx}
\usepackage{booktabs}
\usepackage{multirow}
\usepackage{tikz}
\usepackage{adjustbox}
\usepackage{colortbl}
\usepackage{xcolor}
\usepackage{array}
\usepackage{siunitx}

\definecolor{otherscolor}{RGB}{255,0,0}  
\definecolor{barriercolor}{RGB}{0,255,0}  
\definecolor{bicyclecolor}{RGB}{0,0,255}  
\definecolor{buscolor}{RGB}{255,255,0}  
\definecolor{carcolor}{RGB}{0,255,255}  
\definecolor{constructionvehcolor}{RGB}{255,0,255}  
\definecolor{motorcyclecolor}{RGB}{128,128,0}  
\definecolor{pedestriancolor}{RGB}{0,128,128}  
\definecolor{trafficconecolor}{RGB}{128,0,128}  
\definecolor{trailercolor}{RGB}{255,128,0}  
\definecolor{truckcolor}{RGB}{0,128,255}  
\definecolor{drivablesurfacecolor}{RGB}{128,0,255}  
\definecolor{otherflatsurfacecolor}{RGB}{128,255,0}  
\definecolor{sidewalkcolor}{RGB}{255,0,128}  
\definecolor{terraincolor}{RGB}{0,255,128}  
\definecolor{manmadecolor}{RGB}{128,128,128}  
\definecolor{vegetationcolor}{RGB}{64,64,64}  

\definecolor{generalobjectcolor}{RGB}{255,0,0}  
\definecolor{vehiclecolor}{RGB}{0,255,255}  
\definecolor{pedestriancolor}{RGB}{0,128,128}  
\definecolor{signcolor}{RGB}{255,255,0}  
\definecolor{cyclistcolor}{RGB}{255,0,255}  
\definecolor{trafficlightcolor}{RGB}{0,128,255}  
\definecolor{polecolor}{RGB}{128,0,0}  
\definecolor{conecolor}{RGB}{128,0,128}  
\definecolor{bicyclecolor}{RGB}{0,128,0}  
\definecolor{motorcyclecolor}{RGB}{128,128,0}   
\definecolor{buildingcolor}{RGB}{128,128,128}   
\definecolor{vegetationcolor}{RGB}{0,255,0}  
\definecolor{trunkcolor}{RGB}{128,64,0}  
\definecolor{roadcolor}{RGB}{128,0,255}   
\definecolor{walkablecolor}{RGB}{255,0,128} 

\newcommand{\coloredsquare}[2]{%
  \textcolor{#1}{\rule{0.5em}{0.5em}}\hspace{0.1em}#2%
}

\title{\LARGE \bf
MinkOcc: Towards real-time label-efficient semantic occupancy prediction 
}

\author{Samuel Sze$^{1}$, Daniele De Martini$^{1}$ and Lars Kunze$^{1,2}$
\thanks{$^{1}$Samuel Sze, Daniele De Martini and Lars Kunze are with the Oxford Robotics Institute, Department of Engineering Science, University of Oxford: \texttt{\small (samuels,lars,daniele)@robots.ox.ac.uk}\newline $^{2}$Lars Kunze is with the Bristol Robotics Laboratory, School of Engineering, University of the West of England.}%
}

\begin{document}

\maketitle
\thispagestyle{empty}
\pagestyle{empty}

\begin{abstract}
Developing 3D semantic occupancy prediction models often relies on dense 3D annotations for supervised learning, a process that is both labor and resource-intensive, underscoring the need for label-efficient or even label-free approaches. To address this, we introduce MinkOcc, a multi-modal 3D semantic occupancy prediction framework for cameras and LiDARs that proposes a two-step semi-supervised training procedure. Here, a small dataset of explicitly 3D annotations warm-starts the training process; then, the supervision is continued by simpler-to-annotate accumulated LiDAR sweeps and images -- semantically labelled through vision foundational models. MinkOcc effectively utilizes these sensor-rich supervisory cues and reduces reliance on manual labeling by 90\% while maintaining competitive accuracy. In addition, the proposed model incorporates information from LiDAR and camera data through early fusion and leverages sparse convolution networks for real-time prediction.  With its efficiency in both supervision and computation, we aim to extend MinkOcc beyond curated datasets, enabling broader real-world deployment of 3D semantic occupancy prediction in autonomous driving.

\end{abstract}

\section{Introduction}

Semantic occupancy prediction is crucial for scene understanding in autonomous vehicles (AVs) \cite{survey}, combining spatial and semantic information to enhance decision-making. By estimating the occupied state of each voxel in 3D space, it generalizes well to irregular objects, diverse vehicle shapes, and complex road structures. To support this, various 3D occupancy datasets have emerged, including Monoscene \cite{monoscene}, SSCBench \cite{sscbench}, Occ3D \cite{occ3d}, and OpenScene \cite{openscene}, built upon established multi-modal self-driving datasets like nuScenes \cite{nuscenes}, Waymo \cite{waymo}, and SemanticKITTI \cite{semantickitti}. These datasets span different sensor configurations, semantic taxonomies, and geographic regions, covering diverse environments such as highways and urban areas. 

\begin{figure}[thpb]
    \centering
    \includegraphics[scale=0.4]{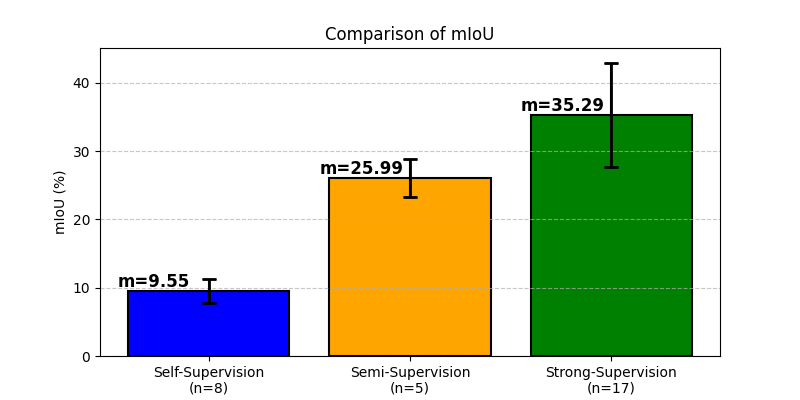} 
    \caption{mIoU of occupancy prediction methods using different supervision signals on Occ3D-nuScenes \cite{occ3d}. Strong-supervision yields the highest mIoU, but its labeling cost is prohibitive. Self-supervision avoids labels but suffers low accuracy. Semi-supervision, where our proposed MinkOcc belongs, offers a practical alternative to reduce labeling costs while maintaining accuracy.}
    \label{miou_comp}
\end{figure}

Figure~\ref{miou_comp} shows that training on these well-curated annotations significantly boosts model performance over weakly-supervised alternatives.
However, large-scale dense annotations across driving scenes remain costly and impractical. For instance, OpenOccupancy \cite{openoccupancy} required approximately 4000 hours of manual labeling for 34{,}000 annotated frames \cite{survey}. This difficulty arises from the need to manually label occluded regions with limited sensor data, and account for dynamic object movements that create spatial-temporal ``tubes'' to ensure temporal consistency of semantic classes across frames \cite{sscbench}. 

While self-supervised learning could eliminate the need for labels, it remains challenging in outdoor environments due to heavy occlusions and limited overlapping viewpoints amid rapid ego-vehicle motion \cite{occnerf, selfocc, renderocc}. Semi-supervised learning offers a practical alternative, markedly reducing annotation dependency while retaining strong supervision where necessary. For example, VFG-SSC \cite{vfgssc} achieves 85\% of its fully supervised performance using only 10\% of labeled data, and similarly facilitates adaptation across diverse AV datasets \cite{survey}.

Building on previous work \cite{minkoccv1}, we introduce MinkOcc, a real-time semi-supervised 3D semantic occupancy prediction model. Our model is built upon Minkowski Engine \cite{choy20194d}, a fully convolutional sparse network serving as the main backbone. For occupancy prediction, we use accumulated LiDAR sweeps as a surrogate for dense 3D ground truth, providing a weak supervision signal for scene completion. For semantic segmentation, we integrate Pulsar \cite{pulsar}, an efficient differentiable neural rendering head that bridges multi-view 2D images with 3D voxel features. Additionally, we employ the vision-language models Grounding-DINO and Segment Anything (SAM) \cite{groundingdino, segmentanything} to generate 2D pseudo-ground-truth labels for semantic supervision. Thanks to its sparse design, MinkOcc achieves real-time inference speeds.

The main contributions of this work are:
\begin{itemize}
    \item A fully sparse, multi-modal 3D semantic occupancy prediction model capable of real-time inference, demonstrating state-of-the-art performance.
    \item A semi-supervised strategy that leverages LiDAR-accumulated sweeps for occupancy completion, as well as differentiable rendering and foundational models for semantic segmentation, thereby reducing reliance on dense and expensive 3D annotations.
\end{itemize}

\begin{figure*}[thpb]
    \centering
        \includegraphics[scale=0.045]{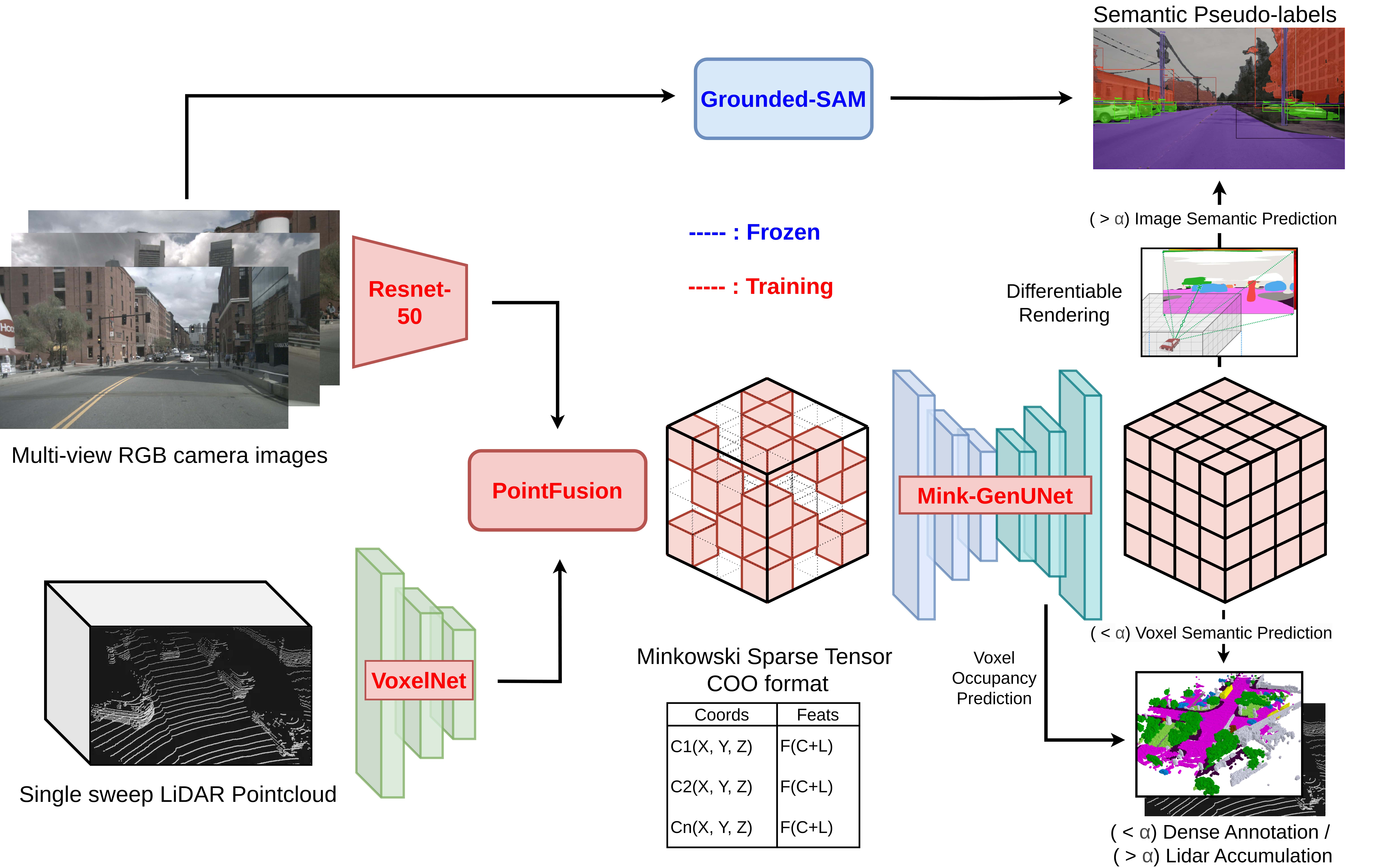} 
    \caption{\textbf{Overview of system pipeline}. Our model predicts dense, 3D semantic occupancy maps from LiDAR and camera information. It is trained in two steps. First, we warm-start the prediction model through \( \alpha = 10 \% \) of dense 3D semantic annotations from Occ3D-nuScenes; then, the voxel semantic prediction branch is turned off, and cheaper LiDAR accumulated sweeps and image semantic maps replace dense annotations. The supervision of the images is provided through a differentiable rendering approach, which projects the semantic information in the camera frames. 3D sparse and dense representations are converted to Coordinate List Format (COO) to ensure compatibility with Minkowski Engine \cite{choy20194d}.}
    \label{system_pipeline}
\end{figure*}

\section{Related Work}

\subsection{Multi-modal occupancy prediction}
Multi-modal 3D semantic occupancy prediction not only leverages richer sensory input but also enhances robustness to sensor failures, such as poor camera visibility at night or reduced LiDAR accuracy in adverse weather \cite{openoccupancy}. In the camera branch, methods typically rely on forward projection via probabilistic depth estimation \cite{lss} or backward projection using either interpolation \cite{simplebev} or cross-attention \cite{bevformer, bevfusion, tpvformer}. In the LiDAR branch, feature extracting backbones such as VoxelNet \cite{voxelnet} and PointPillars \cite{pointpillars} process point clouds into structured representations for downstream processing.

Despite the upsides of multi-modality, \cite{survey} reports that LiDAR-centric methods outperform multi-modal or camera-centric approaches in SemanticKITTI \cite{semantickitti} and SSCBench \cite{sscbench}. Indeed, while LiDAR offers the fastest and most straightforward way for accurate depth, it has often been overlooked in favour for more sophisticated attention methods. Many multi-modal methods predefine 3D voxelized coordinate grids \cite{occfusion, mingoccfusion, nocalibocc, daocc} to query camera features, but these grids frequently include occluded or out-of-view regions, leading to wasted computation. Some works address this by densifying LiDAR points \cite{mrocc} or aggregating LiDAR sweeps over time \cite{occloff} before sensor fusion. In contrast, our approach drastically simplifies the problem by densifying points after fusion, leveraging Minkowski Engine's generative capabilities \cite{generative_me} to handle the complexity of scene completion. As such, sensor fusion is performed through a simple projection transformation of LiDAR points, followed by sampling on the image plane to extract multi-scale camera features, aggregating to a sparse multi-modal feature volume.

\subsection{Semi-supervision in occupancy prediction}

Recent trends in semi-supervision use 2D semantic and depth images as ground truth or derive sparse ground truth by projecting LiDAR semantic segmentation labels onto images \cite{vampire, gsrender, renderocc}. First of all, if LiDAR semantic segmentation ground truth is available, it may be more advantageous to utilize it directly in 3D space and create sparse semantic annotations by accumulating sweeps. In terms of 2D annotation ground truth, curating them across tens of thousands of frames still remains labor-intensive and time-consuming. Therefore, we leverage vision-language models \cite{groundingdino, segmentanything} to rapidly generate 2D pseudo-labels, each assigned a confidence score that reflects the model’s certainty, allowing us to weigh labels accordingly and mitigate errors.


Drawing inspiration from VAMPIRE \cite{vampire}, we aim to construct a robust 3D feature representation of an AV driving scene in a label-efficient manner, which means it is important to have strong 3D supervision signals during the initial learning stage. Conventional semi-supervised methods that exploit 3D labels typically follow a self-training protocol \cite{soccdpt, vfgssc}, where a model is first trained on a small labeled subset, then used to generate pseudo-labels for a larger unlabeled dataset, and subsequently refined using both the original and pseudo-labeled data. However, this approach requires three training phases and is also prone to model overfitting on the small dataset. Instead, we propose integrating weaker 2D pseudo-label signals into the framework to consolidate this multi-step process into a single training phase. In the initial warm-start phase, we leverage dense 3D annotations to guide learning. Once the warm-start iterations are completed, we drop the 3D semantic signal and continue training with 2D supervision.

\subsection{Differentiable neural rendering}

Differentiable rendering enables end-to-end learning of 3D scene representations from 2D image observations. It typically comprises of three stages: a 3D scene representation (e.g., NeRF \cite{nerf}, Gaussian Splatting \cite{3dgs}, or occupancy-based models \cite{3drepresentationsurvey}), a projection method that maps 3D data to 2D via volume rendering, point splatting, or mesh rasterization, and neural shading that refines the 2D output. Many weakly supervised 3D semantic occupancy prediction approaches actively incorporate differentiable rendering \cite{occnerf, renderocc, gausstr, gsrender, selfocc, gaussianformer, gaussianformer2}. RenderOcc \cite{renderocc} introduces a signed density field (SDF) with temporal auxiliary rays to improve multi-view consistency, while OccNeRF \cite{occnerf} extends NeRF with parameterized occupancy fields to render depth and semantic images. GSrender \cite{gsrender} and GaussTR \cite{gausstr} utilize Gaussian splatting for its real-time rendering capabilities.

However, NeRF-based methods rely on volumetric representations that, while enabling fine-grained scene reconstruction, are computationally expensive and require dense multi-view supervision, limiting their practicality in AV scenarios with sparse viewpoints. Gaussian splatting depends on anisotropic covariance matrices for shape adaptation, making optimization difficult in AV environments with only 2-6 views per scene. To address these limitations, we adopt Pulsar \cite{pulsar}, a spherical differentiable renderer optimized for high-speed, CUDA-accelerated rendering. By representing 3D scenes as isotropic spheres, Pulsar enables efficient, geometrically consistent rendering of voxel grids while seamlessly integrating with PyTorch as a fully differentiable component.

\section{Problem Formulation}

We tackle 3D semantic occupancy prediction using six surround-view cameras and a top-mounted LiDAR sensor. The inputs include multi-view images \( I = \{ I_i \}_{i=1}^{N} \), where each \( I_i \in \mathbb{R}^{W \times H \times 3} \) represents an image from the \( i \)-th camera, and a LiDAR point cloud \( P \in \mathbb{R}^{N \times (3 + f)} \) containing 3D coordinates and additional features. The output is a voxel grid \( V \in \mathbb{R}^{X \times Y \times Z} \), where each voxel \( v_{xyz} \) is assigned a class label from \( C \) semantic categories. We adopt a semi-supervised approach, leveraging pseudo-labels from (1) 2D semantic segmentation of current-frame images, \( y_{2D} = \{ y_{2D,i} \}_{i=1}^{N} \), and (2) accumulated LiDAR sweeps over the past \( K \) frames, \( P_{\text{acc}} = \{ P_t \}_{t=-K}^{0} \). 

\section{Method}

\subsection{System Pipeline and Training Procedure}

Figure \ref{system_pipeline} shows the system pipeline. We extract 2D features from multi-view images \(I_i\) using ResNet-50~\cite{resnet} and process LiDAR \(P \in \mathbb{R}^{N \times (3+f)}\) via MVX-Net~\cite{mvxnet}, an extension of VoxelNet~\cite{voxelnet}. We utilize PointFusion~\cite{pointfusion}'s dense fusion to project LiDAR points as spatial anchors onto corresponding image planes. Multi-scale image features are then sampled and adaptively fused across multiple views. The final representation is refined through a shared-weight MLP, producing a sparse yet contextually rich voxelized 3D feature grid.

Following Minkowski Engine's coordinate list format, we convert the 3D feature grid into sparse tensors to be passed into Mink-GenUNet~\cite{generative_me}, a Generative UNet \cite{unet} which performs sparse-to-dense occupancy completion. Pruning is applied at each upsampling layer to remove newly created coordinates using a probability function guided by binary classification against ground truth. Similar to UNet~\cite{unet} architectures, we add skip connections after each generative upsampling to integrate feature details from the encoder. 

Each voxel in the densified 3D feature grid is then classified through a softmax operation into one of the $C$ semantic categories.
We train our system in two separate phases requiring different annotation types.
First, we warm-start our training using a small dataset of dense 3D annotations.
These will directly supervise the 3D occupancy and semantic classification.
In this phase, we denote with $\alpha$ the amount of ground truth data used as a percentage of the total annotated dataset.
After this $\alpha$ warm-start, we switch to accumulated LiDAR sweeps \( P_{\text{acc}} \) for occupancy classification, which are cheaper to annotate but may contain incongruence due to misalignments or moving objects.
For semantic classification, a differentiable rendering head performs 3D-to-2D rasterization at each camera position. We convert voxel features into points at their centers and model them as spheres with learnable radii and opacity. These spheres are then rendered into the 2D image space, where each pixel accumulates contributions from a predefined number of spheres in a back-to-front compositing order. This results in a 2D image feature, which can then be optimized using 2D pseudo labels \(\ y_{2D} \) generated from Gounding-Dino \cite{groundingdino} and SAM \cite{segmentanything}. 

\subsection{Minkowski Convolution Blocks}

Minkowski Engine employs a generalized convolution \cite{choy20194d} as described in Equation \ref{mink_equation}, where \( x_{\text{out}}^u \) represents the output feature at point \( u \).

\begin{equation}
x_{\text{out}}^u = \sum_{i \in \mathcal{N}^D(u, C_{\text{in}})} W_i \cdot x_{\text{in}}^{u + i} \quad \text{for} \quad u \in C_{\text{out}}
\label{mink_equation}
\end{equation}

The neighborhood of input coordinates around \( u \) is defined by \( \mathcal{N}^D(u, C_{\text{in}}) \), which specifies the set of offsets contributing to the sparse convolution operation. The convolution is governed by the learnable weights \( W_i \) of the kernel, while \( x_{\text{in}}^{u + i} \) represents the input feature at the position offset by \( i \) from \( u \). Generalized convolution allows for arbitrary kernel shapes, enabling the direct adoption of 2D architectural design into higher-dimensional networks. Mink-GenUNet utilize two Minkowski convolution layers, each followed by batch normalization and ReLU activation. A residual connection adds the input back to the output after the second convolution. We also insert Minkowski Squeeze and Excite (SE) modules to recalibrate channel-wise features through global average pooling, enhancing focus on key features across sparse data \cite{squeezeandexcite}. To perform sparse-to-dense completion, Minkowski generative transposed convolution blocks are used to expand the sparse tensor's support region. This is determined by the outer product of the convolution kernel applied to the input sparsity pattern, expressed as: \(\text{support}(T) = C \otimes [-K, \ldots, K]^3 \) where \( C \) is the input sparsity pattern and \( K \) is the kernel size.

\subsection{Spherical Differentiable Rendering}

3D voxels are represented as spheres \( S \), where each sphere is parameterized by its voxel center in 3D Cartesian coordinates \( v_{xyz} \), Mink-GenUNet feature vector \( f_{xyz} \in \mathbb{R}^d \), radius \( r_{xyz} \in \mathbb{R} \), and opacity \( o_{xyz} \in \mathbb{R} \). Each scene is learned from a set of training images \( I = \{ I_i \}_{i=1}^{N} \) using their corresponding camera parameters, including rotation \( R_i \), translation \( t_i \), and intrinsic matrix \( K_i \). Given these inputs, spherical differentiable rendering performs differentiable projection \( Proj \) and neural shading \( NS \).

The projection step computes a blending function \(w_{xyz} \) that determines how spheres contribute to each pixel based on their depth ordering, spatial proximity, and opacity. This is calculated by  

\begin{equation}
w_{xyz} = \frac{o_{xyz} \cdot d_{xyz} \cdot \exp(o_{xyz} \cdot z_{xyz} \gamma)}{\exp(\varepsilon \gamma) + \sum_k o_k \cdot d_k \cdot \exp(o_k \cdot z_k \gamma)},
\label{pulsar_weight}
\end{equation}

where \( z_{xyz} \) represents the normalized depth of the sphere, \( d_{xyz} = \min(1, ||\mathbf{d}_{xyz}||_2 / r_{xyz}) \) accounts for the orthogonal distance of the ray to the sphere center, and \( \gamma \) controls the depth sensitivity of the weighting function. Small values of \( \gamma \) enforce sharper blending, while larger values allow for a smoother transition between overlapping spheres. The additional term \( \varepsilon \) ensures numerical stability, preventing division errors. Equation \ref{pulsar_weight} also ensures that spheres closer to the camera and with higher opacity contribute more significantly to the final pixel value. 

After projection, we aggregate the weighted contributions from all intersecting spheres on each pixel. Given the per-pixel blending weights \( w_{xyz} \), the rendered feature value at each pixel is computed as a weighted sum: \( F_{uv} = \sum w_{xyz} f_{xyz}. \). The shading function \( NS \) with its parameters \( \Theta \) then transforms the accumulated features \( F \), which encapsulates all camera viewpoints, into semantic prediction \( \hat{{y_{2D}}} = NS(F; \Theta) \).  \( \Theta \) is essentially a  convolutional UNet which performs semantic segmentation in our case.

Altogether, we form Equation \ref{pulsar_complete}. The overall optimization objective is to minimize the difference between predicted semantics \( \hat{y_{2D}} \) and ground-truth pseudo-labels \( y_{2D} \). As such, we refine both the spheres \( S \) and neural shader parameters \( \Theta \) to reach their optimized levels \(S^*, \Theta^*\).

\begin{equation}
\begin{split}
S^*, \Theta^* \quad & = \\ \arg \min_{S, \Theta} \sum_{i}  
& \left| y_{2D,i} - NS \left( Proj(S, R_i, t_i, K_i); \Theta \right) \right|
\end{split}
\label{pulsar_complete}
\end{equation}

\subsection{Loss Function}

We use three losses to supervise our training. \( \mathcal{L}_{\text{bce}} \) is a voxel-based binary cross-entropy loss applied for sparse-to-dense occupancy completion across the decoder layers. We adopt \( \mathcal{L}_{\text{3D\_ce}} \) as a dense 3D class-balanced cross-entropy loss against 3D annotation. Class balancing reweights each semantic category based on its frequency, calculated using an effective number of voxel samples per class with \( \beta = 0.9 \) over the entire dataset \cite{42dot}.

\begin{equation}
\mathcal{L}_{\text{3D\_ce}}(\hat{y}, y) = - \frac{1 - \beta}{1 - \beta^{n_y}} \log \left( \frac{\exp(\hat{y}_y)}{\sum_{j=1}^{C} \exp(\hat{y}_j)} \right).
\end{equation}

2D pseudo-labels are not perfect and can be prone to errors. Therefore, we use a soft cross-entropy loss weighted by the confidence score from Grounded-DINO's object detection. For overlapping detections on the same image pixel, we select the highest confidence score among them. The soft cross-entropy loss is given by:

\begin{equation}
\mathcal{L}_{\text{2D\_ce}} = - \sum_{i \in \mathcal{I}} c_i \cdot y_{\text{2D},i} \log \hat{y}_{\text{2D},i},
\label{sce_loss_eqn}
\end{equation}

where \( y_{\text{2D},i} \) is the one-hot pseudo-label from Grounded-DINO and SAM, \( \hat{y}_{\text{2D},i} \) is the predicted softmax probability for pixel \( i \), and \( c_i \) is the confidence score assigned to each pseudo-label. Unlabeled pixels (class 0) are ignored. This formulation allows high-confidence labels to contribute more significantly to the loss while low-confidence labels contribute less.

The total loss is shown in Equation \ref{total_loss_eqn}, balanced by their respective hyperparameters \( \lambda_1 \) and \( \lambda_2 \):

\begin{equation}
\mathcal{L}_{\text{total}} =
\begin{cases} 
\lambda_1 \mathcal{L}_{\text{3D\_ce}} + \lambda_2 \mathcal{L}_{\text{bce}}, & \text{if }  \text{ warm-start phase} \\
\lambda_1 \mathcal{L}_{\text{2D\_ce}} + \lambda_2 \mathcal{L}_{\text{bce}}, & \text{if }  \text{ after warm-start}
\end{cases}
\label{total_loss_eqn}
\end{equation}
In our training, the warm-start phase employs dense 3D annotations for occupancy completion and semantic segmentation using \( \mathcal{L}_{\text{bce}} \) and \( \mathcal{L}_{\text{3D\_ce}} \), respectively. After the warm-start, semantic supervision transitions to 2D pseudo-labels via \( \mathcal{L}_{\text{2D\_ce}} \), while occupancy completion continues with LiDAR accumulation pseudo ground truth using \( \mathcal{L}_{\text{bce}} \). \( \lambda_1 \) and \( \lambda_2 \) are empirically set, with \( \lambda_1 = 0.5 \) and \( \lambda_2 = 1 \).

\begin{table*}[htbp]
\vspace{4pt}
\caption{Quantitative results on Occ3D-nuScenes validation dataset. Self-supervision mIoU excludes \textit{others} and \textit{other flat} as they are not well defined in 2D labels \cite{occnerf}. Best results for each semantic class are bolded.}
\centering
\resizebox{\textwidth}{!}{%
\setlength{\tabcolsep}{2pt}
\begin{tabular}{
  l
  c
  S[table-format=2.2]
  S[table-format=2.2]
  *{17}{S[table-format=2.2]}
}
\toprule
\textbf{Method} & \textbf{Sup.} & {\textbf{RayIoU}} & {\textbf{mIoU}} & 
{\rotatebox{90}{\coloredsquare{otherscolor}{others}}} & 
{\rotatebox{90}{\coloredsquare{barriercolor}{barrier}}} & 
{\rotatebox{90}{\coloredsquare{bicyclecolor}{bicycle}}} & 
{\rotatebox{90}{\coloredsquare{buscolor}{bus}}} & 
{\rotatebox{90}{\coloredsquare{carcolor}{car}}} & 
{\rotatebox{90}{\coloredsquare{constructionvehcolor}{const. veh.}}} & 
{\rotatebox{90}{\coloredsquare{motorcyclecolor}{motorcycle}}} & 
{\rotatebox{90}{\coloredsquare{pedestriancolor}{pedestrian}}} & 
{\rotatebox{90}{\coloredsquare{trafficconecolor}{traffic cone}}} & 
{\rotatebox{90}{\coloredsquare{trailercolor}{trailer}}} & 
{\rotatebox{90}{\coloredsquare{truckcolor}{truck}}} & 
{\rotatebox{90}{\coloredsquare{drivablesurfacecolor}{driv. surf.}}} & 
{\rotatebox{90}{\coloredsquare{otherflatsurfacecolor}{other flat}}} & 
{\rotatebox{90}{\coloredsquare{sidewalkcolor}{sidewalk}}} & 
{\rotatebox{90}{\coloredsquare{terraincolor}{terrain}}} & 
{\rotatebox{90}{\coloredsquare{manmadecolor}{manmade}}} & 
{\rotatebox{90}{\coloredsquare{vegetationcolor}{vegetation}}} \\
\midrule


OccNeRF \cite{occnerf} & Self & \text{-}  & 10.81 & \text{-}  & \textbf{0.83} & \textbf{0.82} & 5.13 & 12.49 & \textbf{3.50} & \textbf{0.23} & \textbf{3.10} & \textbf{1.84} & \textbf{0.52} & \textbf{3.90} & 52.62 & \text{-}  & 20.81 & 24.75 & 18.45 & 13.19 \\
SelfOcc \cite{selfocc} & Self & \text{-}  & 9.03 & \text{-}  & 0.00 & 0.00 & 0.00 & 10.03 & 0.00 & 0.00 & 0.00 & 0.00 & 0.00 & \textbf{7.11} & 52.96 & \text{-}  & \textbf{23.59} & \textbf{25.16} & 11.97 & 4.61 \\

\textbf{MinkOcc} & Self & \textbf{12.5} & \textbf{13.23} & \text{-} & 0.00 & 0.00 & \textbf{7.92} & \textbf{13.83} & 0.00 & 0.00 & 0.00 & 0.00 & 0.00 & 5.75 & \textbf{70.27} & \text{-} & 17.60 & 20.08 & \textbf{35.97} & \textbf{26.98} \\

\midrule

RenderOcc \cite{renderocc} & Semi & 19.50 & 23.93 & 5.69 & 27.56 & 14.36 & 19.91 & 20.56 & 11.96 & 12.42 & 12.14 & 14.34 & 20.81 & 18.94 & 68.85 & 33.35 & 42.01 & 43.94 & 17.36 & 22.61 \\
VAMPIRE \cite{vampire} & Semi & \text{-} & 28.33 & 7.48 & 32.64 & \textbf{16.15} & \textbf{36.73} & 41.44 & 16.59 & 20.64 & 16.55 & 15.09 & 21.02 & 28.47 & 67.96 & 33.73 & 41.61 & 40.76 & 24.53 & 20.26 \\
GSRender \cite{gsrender} & Semi & \text{-} & 29.56 & \textbf{8.42} & 34.93 & 15.12 & 30.71 & 29.61 & 16.70 & 9.48 & 17.61 & \textbf{16.58} & \textbf{23.92} & 27.24 & \textbf{77.94} & \textbf{39.21} & \textbf{51.69} & \textbf{54.61} & 23.82 & 24.92 \\

\textbf{MinkOcc} & Semi & \textbf{24.60} & \textbf{33.43} & 1.58 & \textbf{37.90} & 10.24 & 34.95 & \textbf{46.20} & \textbf{22.08} & \textbf{35.07} & \textbf{37.46} & 15.06 & 15.59 & \textbf{35.57} & 73.13 & 37.01 & 46.84 & 48.23 & \textbf{40.61} & \textbf{30.75} \\

\midrule
Occ3D \cite{occ3d} & Strong & \text{-}  & 28.53 & 8.09 & 39.33 & 20.56 & 38.29 & 42.24 & 16.93 & 24.52 & 22.72 & 21.05 & 22.98 & 31.11 & 53.33 & 33.84 & 37.98 & 33.23 & 20.79 & 18.00 \\
FastOcc \cite{fastocc} & Strong & \text{-}  & 40.75 & 12.86 & 46.58 & 29.93 & 46.07 & 54.09 & 23.74 & 31.10 & 30.68 & 28.52 & 33.08 & 39.69 & 83.33 & 44.65 & 53.90 & 55.46 & 42.61 & 36.50 \\
FB-OCC \cite{nvocc} & Strong & 33.5 & 42.06 & 14.30 & 49.71 & \textbf{30.00} & 46.62 & 51.54 & \textbf{29.30} & 29.13 & 29.35 & \textbf{30.48} & 34.97 & 39.36 & 83.07 & \textbf{47.16}& \textbf{55.62} & \textbf{59.88} & 44.89 & 39.58 \\

\textbf{MinkOcc} & Strong & \textbf{40.90} & \textbf{44.85 } & \textbf{20.02} & \textbf{56.45} & 25.79 & \textbf{55.60} & \textbf{57.13} & 27.54 & \textbf{40.22} &\textbf{39.15} & 29.59 & \textbf{41.39} & \textbf{46.29} & \textbf{85.78} & 46.84 & 45.43 & 46.80 & \textbf{50.42} & \textbf{48.03} \\

\bottomrule
\end{tabular}
}
\label{tab:occ3dnuscenes_table}
\end{table*}


\begin{figure}[thpb]
    \centering
    \includegraphics[scale=0.14]{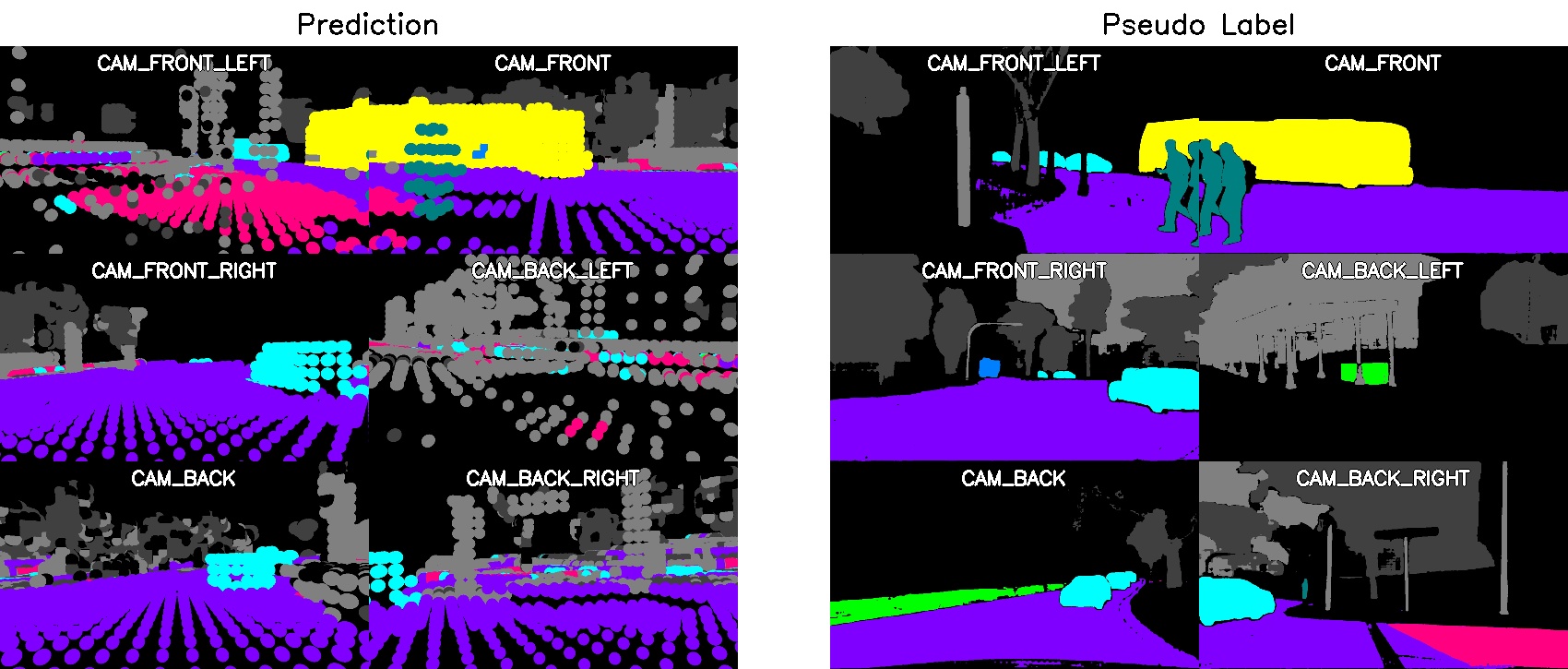} 
    \caption{Comparison of MinkOcc-semi's predicted 3D feature volume rendered in 2D against 2D pseudo-label. Results are from Occ3D-nuScenes validation set. \textbf{Better viewed in color and zoomed in}.}
    \label{2D_vis}
\end{figure}

\begin{figure*}[thpb]
    \centering
    \includegraphics[scale=0.04]{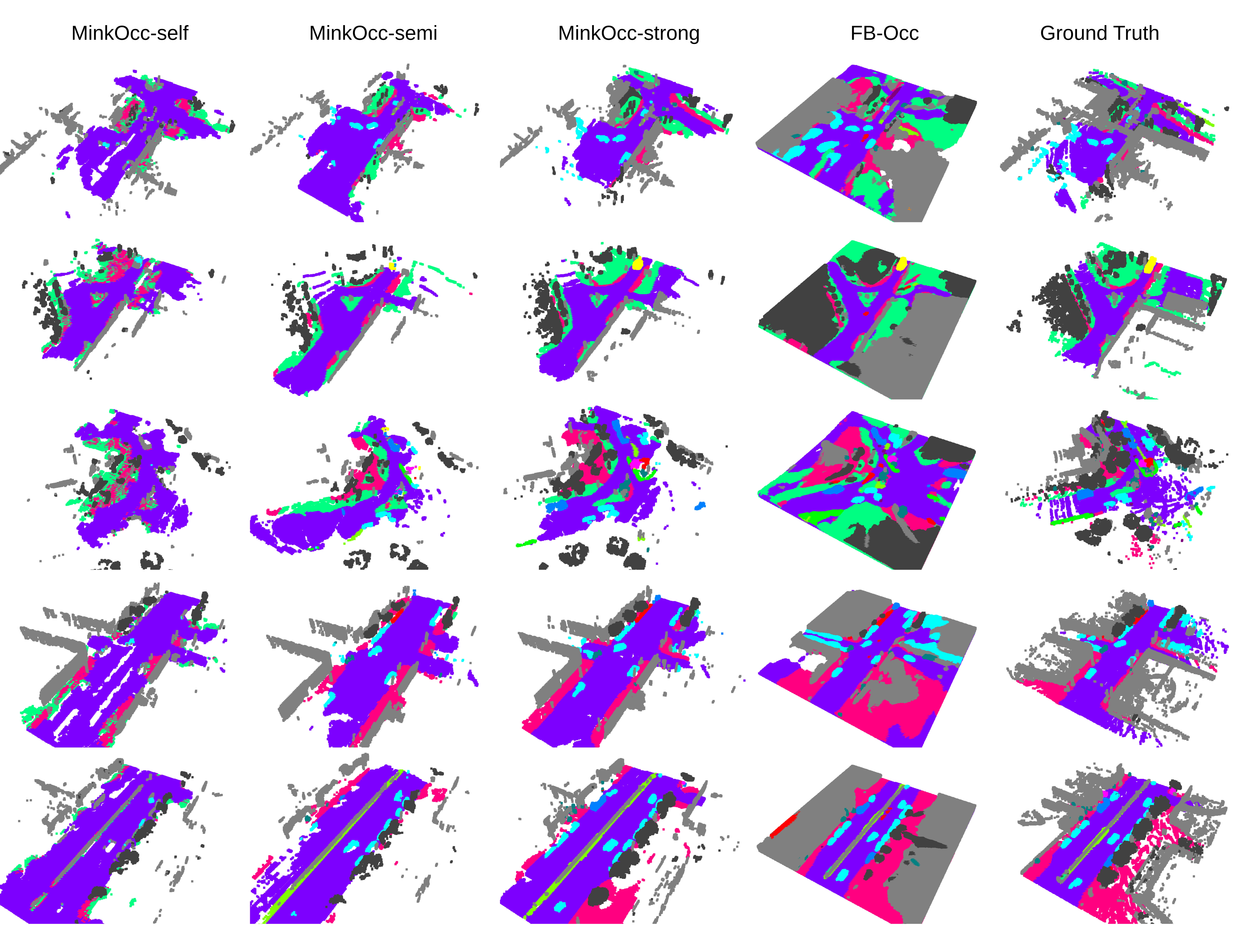} 
    \caption{Qualitative results on Occ3D-nuScenes validation across different MinkOcc warm-start phase percentage setup against FB-Occ \cite{nvocc} and Ground Truth. \textbf{Better viewed in color and zoomed in.}}
    \label{qualitative}
\end{figure*}

\section{Experimental Setup}

\subsection{Datasets}

We conduct our experiments on the Occ3D-nuScenes \cite{occ3d} dataset, which consists of 28,130 training scenes and 6,019 validation scenes. Each scene includes six RGB cameras forming a 360-degree surround view, calibrated LiDAR scans and voxelized 3D ground truth. The voxel grid has dimensions \(200 \times 200 \times 16\), covering a physical space from \([-40\,\text{m}, -40\,\text{m}, -1\,\text{m}]\) to \([40\,\text{m}, 40\,\text{m}, 5.4\,\text{m}]\), with 18 semantic classes, including free space (class 17). We use the masks provided to occlude camera out-of-view voxels. For semi-supervised training, we set warm-start phase as \(\alpha = 10\%\) of total scenes from Occ3D-nuScene. We use Grounding-DINO~\cite{groundingdino} to generate bounding box proposals from text prompts, followed by SAM ~\cite{segmentanything} for per-pixel semantic segmentation, producing 2D pseudo labels. We select semantic classes that align better with 2D representations to improve segmentation quality in perspective space. Since our primary focus is 3D semantic occupancy, we subsequently remap the 2D labels to match the 3D annotations in Occ3D-nuScenes, ensuring consistency across tasks.

\subsection{Training details}
Training is configured with the Adam optimizer at a learning rate of 1e-4, and a Cosine-Annealing learning rate schedule. The model is trained for 30 epochs, with a batch size of 2.  The hardware setup for the training included 1 NVIDIA RTX 4090. Data augmentation is performed on the camera, LiDAR and annotated 3D ground truth. For 2D data augmentation, we perform random image translation, scaling, rotation and RGB perturbation. For 3D data augmentation, we perform random flipping, rotation about the z-axis and ground truth voxel dropout. 
We also perform gradient clipping during warm-start phase $\alpha$ transition to stabilize training.

\subsection{Evaluation Metrics}

We evaluate our 3D semantic occupancy prediction using the traditional Mean Intersection over Union (mIoU) for voxel-level metrics as well as the Ray-level metric (RayIoU) proposed by SparseOcc \cite{sparseocc}. mIoU is defined as \(\frac{1}{|\underline{C}|} \sum_{i \in \underline{C}} \frac{TP_i}{TP_i + FP_i + FN_i} \) where \( \underline{C} \) represents the semantic classes excluding the ``free'' class, and \( TP_i \), \( FP_i \), and \( FN_i \) denote True Positives, False Positives, and False Negatives for class \( i \). However, imperfections in dense annotations can cause challenges in evaluating physically thin objects and occlusion. This results in disproportionately strict voxel-level metrics, where even a one-voxel deviation may reduce the IoU to zero. As such, to address this limitation, we also use RayIoU, which simulates a LiDAR beam by using query rays to sample the first voxel contact surface.

\section{Experiments}

\subsection{Semi-supervised Semantic Occupancy Prediction}

Referring to Table \ref{tab:occ3dnuscenes_table}, we compare our model's performance with state-of-the-art semi-supervised methods \cite{renderocc, vampire, gsrender}. With a warm-start phase of $\alpha = 10\%$ on the Occ3D-nuScenes dataset, our model demonstrates strong performance across key dynamic object categories. It achieves competitive results in \textit{cars}, \textit{motorcycles}, \textit{pedestrians}, and \textit{trucks}, demonstrating the effectiveness of 2D supervision in compensating for limited 3D labels. Figure \ref{2D_vis} compares our rendered 2D predictions with pseudo-labels from the Occ3D-nuScenes validation set, illustrating how the model refines semantic segmentation in perspective space. By training with 2D pseudo-labels from vision-language models, the model learns a semantically meaningful 3D feature representation, improving both 2D segmentation accuracy and its 3D semantic occupancy predictions.

Performance in other classes is slightly weaker, likely due to suboptimal pseudo-label generation. For example, in Figure \ref{2D_vis}, certain pseudo-label frames fail to annotate regions of \textit{drivable surface} and \textit{sidewalk}. Errors in pseudo-labels, which is reflected by the soft cross-entropy 2D loss, weakens supervision for these classes. Addressing this may require improving pseudo-label generation or refining the model’s handling of uncertain supervision. 

As shown in Figure \ref{qualitative}, our method maintains strong geometrical accuracy, ensuring precise voxel alignment with real-world positions. This is evident in dense urban traffic scenes such as row 1, row 3, and row 4, where \textit{cars}, \textit{pedestrians}, and \textit{trucks} retain their correct shapes instead of appearing stretched or merged, as seen in FB-Occ. The high RayIoU score further indicates that the model reconstructs 3D structures without hallucinating excessive thickness behind occlusions, making it more reliable in cluttered environments.

In semi-supervised settings, sparsity plays a crucial role in scene densification and semantic accuracy. When voxelized, LiDAR accumulation provides fewer points than dense 3D annotations, and at coarser voxel resolutions, multiple LiDAR points may merge into a single voxel. Additionally, the voxel-to-point-to-sphere transformation introduces information loss by collapsing each voxel into a single point before rendering, discarding some semantic details. Referring to Figure \ref{2D_vis}, these factors result in fewer rendered spheres for \textit{pedestrians} and distant \textit{cars}, limiting the model’s ability to align predictions with pseudo-labels. To better understand these effects, we conduct an ablation study on voxel resolution and its impact on accuracy and inference speed. Further research is also needed to improve voxel-to-sphere transformation pipeline to avoid unnecessary information loss.

In terms of inference speed, referring to Table \ref{tab:voxel_resolution_alpha}, we achieve a potential inference speed of 44.69 milliseconds (ms) on MinkOcc-semi, representing a \textit{2}-times increase compared to the TensorRT-optimized FastOcc \cite{fastocc} which runs at 80 ms. This translates to 23 FPS while utilizing only 1.65 GB of GPU memory. Our video demo further demonstrates the model's capability of real-time inference.

\subsection{Ablation Study}


\begin{table}[htbp]
\caption{Effect of Voxel Resolution and $\alpha$ on Model Performance}
\centering
\resizebox{0.45\textwidth}{!}{%
\setlength{\tabcolsep}{3.5pt} 
\begin{tabular}{
  c  
  c  
  c  
  c  
}
\toprule
\textbf{Voxel Resolution (m)} & {\boldmath$\alpha$ (\%)} & {\textbf{Latency (ms)}} & {\textbf{mIoU} (\%)} \\
\midrule
0.4 & 100 & 43.18 & 44.85 \\
0.4 & 10  & 44.69 & 33.43 \\
0.4 & 0   & 42.47 & 13.23 \\
\midrule

0.2 & 10  & 54.99 & 36.23 \\
\midrule

0.1 & 10  & 102.31 & 37.05 \\
\bottomrule
\end{tabular}
}
\label{tab:voxel_resolution_alpha}
\end{table}


To evaluate the impact of dense 3D supervision, we vary warm-start phase $\alpha$ which controls the fraction of training samples with dense 3D annotations. At $\alpha = 0 \%$, MinkOcc operates in a fully self-supervised setting, relying solely on 2D pseudo-labels and LiDAR-accumulated scans. While this enables open-vocabulary classification via Grounded-SAM, we remap 2D labels back to 3D (Section V.B) for fair comparison with other methods. As shown in Table \ref{tab:occ3dnuscenes_table} and Figure \ref{qualitative}, the model effectively segments larger road-layout-specific classes such as \textit{drivable surface}, \textit{manmade}, \textit{sidewalk}, and \textit{vegetation}. However, it struggles with smaller, less frequent objects, indicating that 2D semantic pseudo-labels alone is insufficient for self-supervised semantic occupancy prediction. Additional supervision from LiDAR, camera signals, or temporal information is needed.

At $\alpha = 100\%$, MinkOcc operates in a strongly supervised mode, with 2D pseudo-labels acting as auxiliary supervision alongside dense 3D annotation. MinkOcc-strong achieves better overall semantic accuracy over Occ3D, FB-Occ, and FastOcc while maintaining efficient inference. Similar to MinkOcc-semi, it reconstructs scenes with high geometric fidelity, avoiding generating artifacts that distort object shapes and road layouts.

We analyze the effect of voxel resolution on accuracy and inference speed. Table \ref{tab:voxel_resolution_alpha} highlights the trade-off between voxel size, latency, and mIoU. While higher resolution (smaller voxels) improves mIoU, the computational cost rises exponentially. The limited accuracy gains at 0.1m suggest that this voxel resolution exceeds the inherent spatial precision of LiDAR sensors in nuScenes \cite{nuscenes}, leading to an occupancy grid of many empty voxels. A moderate upsample in voxel resolution (e.g., from 0.4m to 0.2m) in a semi-supervised setting provides a good balance, improving performance with minimal impact on efficiency.

\section{Conclusion}

In this work, we introduced MinkOcc, a real-time, label-efficient 3D semantic occupancy prediction model that leverages semi-supervision to reduce reliance on dense 3D annotations. Our approach effectively balances accuracy and computational efficiency by integrating sparse convolution networks with multi-modal sensor fusion and differentiable rendering. Experimental results on the Occ3D-nuScenes dataset demonstrate that MinkOcc-semi achieves competitive performance with significantly fewer annotations, particularly excelling in dynamic object segmentation while maintaining real-time inference speeds.

Our ablation study also examines voxel resolution, supervision quality, and computational cost in 3D semantic occupancy prediction. In our main experiments, we adopt a voxel resolution of 0.4 m to align with the resolution provided by dense 3D annotations. Notably, a finer resolution of 0.2 m yields better performance, suggesting that capturing finer spatial details can be advantageous under weak supervision. However, the benefits of higher voxel resolution are limited by the inherent sparsity of LiDAR-accumulated sweeps. Overall, while strongly supervised methods continue to set the benchmark, our results indicate that semi-supervised approaches hold promise for AV datasets with limited dense annotations.

Future work could improve 2D supervision quality by enhancing semantic pseudo-label accuracy through depth and image feature maps. Another key direction is to develop a denser rendering approach for voxel grids, either via direct voxel rendering \cite{voxelrendering} or a lossless voxel-to-sphere transformation, to mitigate information loss. Another direction would be applying MinkOcc to diverse AV datasets to further validate its robustness and generalization.

\addtolength{\textheight}{0cm}   




\section*{ACKNOWLEDGMENT}
This project was supported by the EPSRC project RAILS (EP/W011344/1) and the Oxford Robotics Institute's research project RobotCycle. 

\bibliographystyle{IEEEtran}
\bibliography{references}


\end{document}